\documentclass[nohyperref]{article}

\usepackage{microtype}
\usepackage{url}
\usepackage{multirow}
\usepackage{array,multirow,graphicx}
\usepackage{subfigure}
\usepackage{amsmath,lipsum}
\usepackage{booktabs}

\usepackage[accepted]{icml2023}

\usepackage{amsmath}
\usepackage{amssymb}
\usepackage{mathtools}
\usepackage{amsthm}

\usepackage[utf8]{inputenc} 
\usepackage[T1]{fontenc}    
\usepackage{amsfonts}       
\usepackage{nicefrac}       
\usepackage{xcolor}         
\usepackage{soul}
\usepackage{amsmath, bm}
\usepackage{dsfont}
\usepackage{amssymb}
\usepackage{multirow}

\usepackage[vlined,linesnumbered,ruled,algo2e]{algorithm2e}
\usepackage[linesnumbered,ruled,vlined]{algorithm2e}
\usepackage[flushleft]{threeparttable}

\usepackage{varwidth}
\usepackage{pifont}
\usepackage{makecell}
\usepackage{wrapfig}
\usepackage{enumitem}
\usepackage{caption}
\usepackage{dsfont}
\usepackage{soul}
\usepackage{varwidth}
\usepackage{pifont}
\usepackage{makecell}
\usepackage{enumitem}
\usepackage[hang,flushmargin]{footmisc}

\usepackage{xspace}

\usepackage{microtype}
\usepackage{graphicx}
\usepackage{booktabs} 
\usepackage{xspace}
\usepackage{colortbl}
\usepackage{array,multirow,graphicx}
\usepackage{makecell}
\usepackage{textcomp}
\usepackage[allbordercolors=blue]{hyperref}
\usepackage{cleveref}
\usepackage{multirow}
\theoremstyle{plain}

\theoremstyle{definition}

\theoremstyle{remark}

\usepackage[textsize=tiny]{todonotes}

\begin{document}

\twocolumn[
\icmltitle{Brainformers: Trading Simplicity for Efficiency}




\icmlsetsymbol{equal}{*}

\begin{icmlauthorlist}
\icmlauthor{Yanqi Zhou}{yyy}
\icmlauthor{Nan Du}{yyy}
\icmlauthor{Yanping Huang}{yyy}
\icmlauthor{Daiyi Peng}{yyy}
\icmlauthor{Chang Lan}{yyy}
\icmlauthor{Da Huang}{yyy}
\icmlauthor{Siamak Shakeri}{yyy}
\icmlauthor{David So}{yyy}
\icmlauthor{Andrew Dai}{yyy}
\icmlauthor{Yifeng Lu}{yyy}
\icmlauthor{Zhifeng Chen}{yyy}
\icmlauthor{Quoc Le}{yyy}
\icmlauthor{Claire Cui}{yyy}
\icmlauthor{James Laudon}{yyy}
\icmlauthor{Jeff Dean}{yyy}
\end{icmlauthorlist}

\icmlaffiliation{yyy}{Google Deepmind}

\icmlcorrespondingauthor{Yanqi Zhou}{yanqiz@google.com}

\icmlkeywords{Machine Learning, ICML}

\vskip 0.3in
]



\printAffiliationsAndNotice{}  

\begin{abstract}
Transformers are central to recent successes in natural language processing and computer vision. Transformers have a mostly uniform backbone where layers alternate between feed-forward and self-attention in order to build a deep network. Here we investigate this design choice and find that more complex blocks that have different permutations of layer primitives can be more efficient. Using this insight, we develop a complex block, named Brainformer, that consists of a diverse sets of layers such as sparsely gated feed-forward layers, dense feed-forward layers, attention layers, and various forms of layer normalization and activation functions. Brainformer consistently outperforms the state-of-the-art dense and sparse Transformers, in terms of both quality and efficiency. A Brainformer model with 8 billion activated parameters per token demonstrates 2$\times$ faster training convergence and 5$\times$ faster step time compared to its GLaM counterpart. In downstream task evaluation, Brainformer also demonstrates a 3\% higher SuperGLUE score with fine-tuning compared to GLaM with a similar number of activated parameters. Finally, Brainformer largely outperforms a Primer dense model derived with NAS with similar computation per token on fewshot evaluations. 
\end{abstract}

\section{Introduction}
\label{sec:intro}

\begin{figure}[ht]
  \centering 
\includegraphics[width=0.98\linewidth,trim={0cm 0cm 0cm, 0cm},clip]{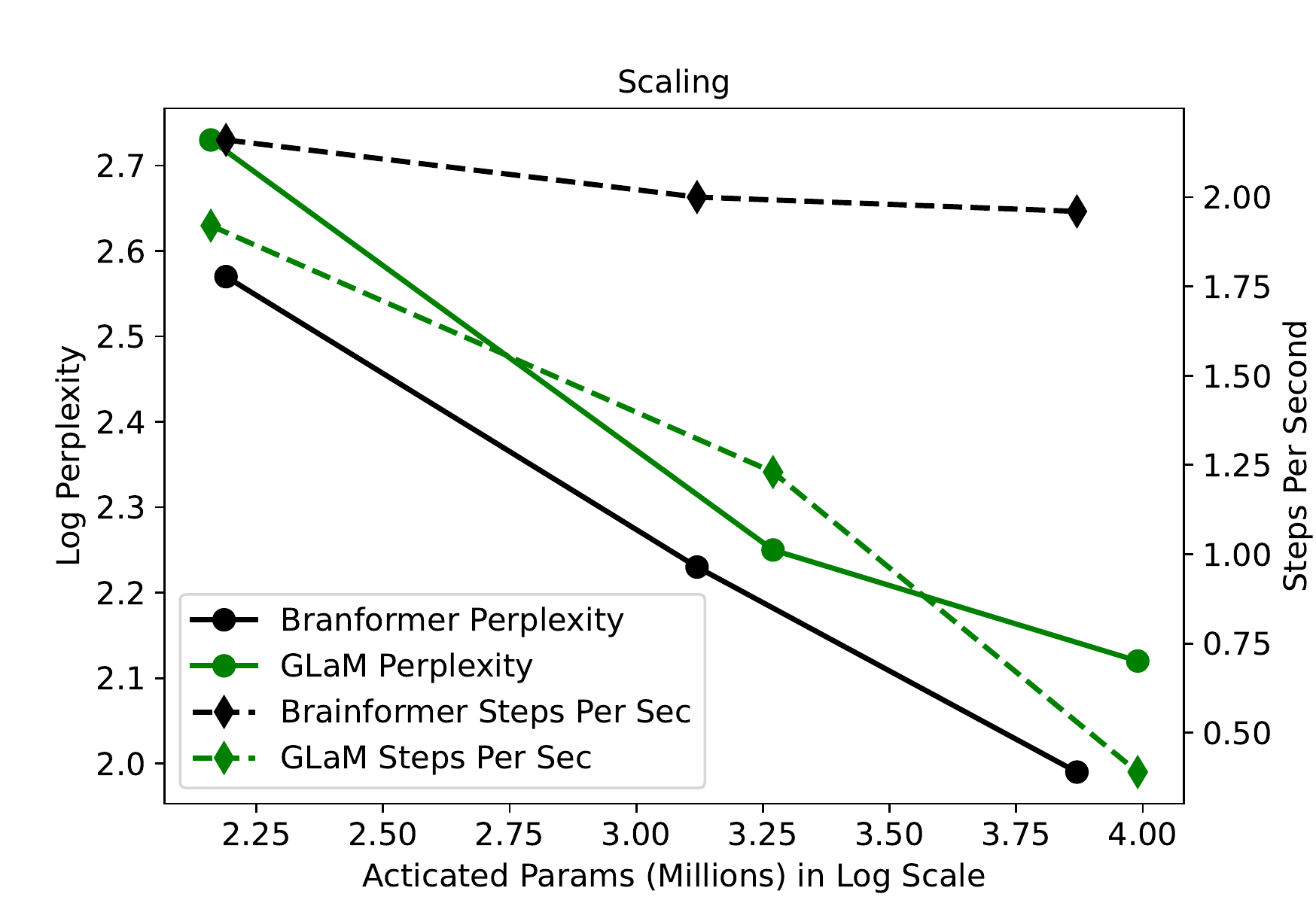}
    \caption{Brainformer Vs. GLaM in Scaling. Brainformer improves model quality at much faster training step time.}
    \label{fig:scaling}
\end{figure}
\begin{figure}[ht]
  \centering 
\includegraphics[width=0.9\linewidth,trim={1.5cm 2cm 1.5cm, 2cm},clip]{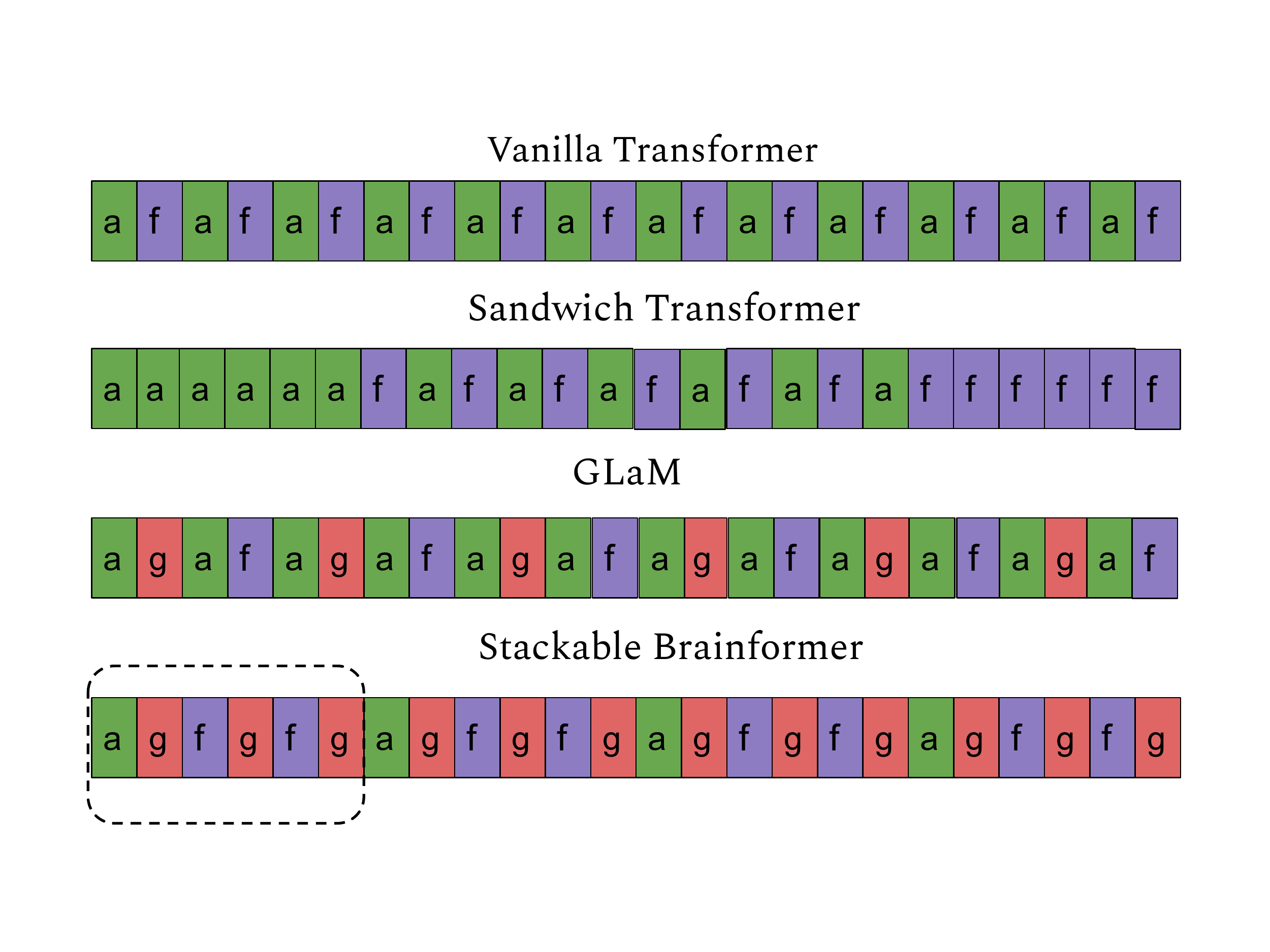}
    \caption{High-level Comparison with Related Work. 'a': attention, 'f': feed-forward, 'g': sparsely gated feed-forward. GLaM interleaves dense transformer blocks with sparse transformer blocks. Brainformer reduces the frequency of attention and changes layer widths together with layer types.}
    \label{fig:high-level}
\end{figure}
In recent years, large neural networks derived from from the Transformer architecture~\citep{vaswani2017attention} have demonstrated superior results on language understanding and generative tasks. Many improvements on Transformer variants have come from scaling the size of models~\citep{raffel2020exploring, NEURIPS2020_gpt3,shoeybi2020megatronlm,chowdhery2022palm}, scaling the training tokens~\citep{hoffmann2022training,shoeybi2020megatronlm}, better training data quality~\citep{du2022glam}, and sparsely activated model architectures~\citep{du2022glam,lepikhin2020gshard,roller2021hash,lewis2021base}.

Among the efficient transformer language models~\citep{wang2020linformer,choromanski2020rethinking,tay2021synthesizer,hua2022transformer}, there is a focus on improving attention-layer efficiency using low-rank approaches or approximations. However, recent work has also identified that dense feed-forward layers constitute most of the computational cost for common sequence lengths ($\leq$2048), particularly when the model is large~\citep{du2022glam,expertchoice2022}. To further improve compute efficiency such as total FLOPs used during training to reach convergence, sparsely gated Mixture-of-Experts ~\citep{lepikhin2020gshard,fedus2021switch,du2022glam,expertchoice2022,roller2021hash,lewis2021base,jaszczur2021sparse} have become prevalent, giving the model a larger overall capacity to improve quality while holding computational cost fixed. Sparsely activated models not only reduce the computational cost, but also have better specialization by training different experts on different data distributions through the use of a routing function without reducing the effective training time for each expert. The MoE architectures in this line of work are based on uniform transformer blocks or interleaving dense and sparse layers~\citep{du2022glam} and a fixed top-k routing.


Resonating with the layer-wise architecture stacking in EfficientNet~\citep{tan2019efficientnet} and layer reordering in the sandwich transformer~\citep{press2019improving}, we propose a non-uniform architecture with sparsity where there is no strict layer interleaving as in the vanilla transformer in \cref{fig:high-level}. We trade off architecture regularity by allowing the search space to compose different sub-layers in different orders. For better scaling, we introduce sparsity in the search space with a sparsely gated feed-forward layer (MoE layer) coupled with different gating mechanisms. 

We find that optimizing the architecture, sparsity, and routing mechanism in sparse layers is critical to achieve near-perfect log-scale scaling in quality. Figure~\ref{fig:scaling} shows that Brainformer scales much better than GLaM (manually crafted sparse transformer). Brainformer consistently improves training perplexity while keeps example rate almost constant when increasing model capacity, however, GLaM has a much worse example rate when scaled up.

We only treat the MoE layer as a general method to sparsify the model. In practice, any conditional computation method can be blended in. We apply a simple evolutionary search to discover many attributes, such as the best way to interleave layers and layer capacities, when to fuse layers, and when to specialize layers with MoE modules. For ease of scaling, we propose a block-wise sub-layer grouping, such that stacking a variable number of blocks produces models of different scales, as illustrated in Stackable Brainformer in \cref{fig:high-level}. As our results in Section~\ref{sec:evaluation} show, this approach has proven effective in our evaluation at multiple model scales.

\section{Related Work}
\label{sec:related}

\textbf{Large Language Models:}
Language models have demonstrated strong performance for many natural language processing tasks~\citep{mikolov2010recurrent,sutskever2011generating,NIPS2015_7137debd}. Scaling up model capacity and number of training tokens has shown huge success in enhancing the performance of computer vision architectures~\citep{he2015deep, he2016identity, ghiasi2019nasfpn, dai2021coatnet} as well as neural language models~\citep{GPT2018, GPT32020, kaplan2020scaling, raffel2020exploring,shoeybi2020megatronlm,hoffmann2022training}.

\textbf{Sparsely Activated Models:}
Conditional computation effectively increases the capacity of a deep neural network without increasing the total amount of computation, by activating certain parameters and computation on demand, based off the input token or sequence~\citep{cho2014exponentially,puigcerver2020scalable, lin2019conditional}. The gating decisions may be binary or sparse and continuous, stochastic or deterministic. In a multi-device setting, sparsely-gated MoE~\citep{shazeer2017outrageously} demonstrates massive improvements in model capacity, training time, or model quality with gating. Various MoE architectures including Switch Transformer~\citep{fedus2021switch} and GLaM~\citep{du2022glam} have been proposed. They adopt a token-based gating where an auxiliary loss is imposed to counter load imbalance issues. Recently, more advanced gating functions are devised to ameliorate load imbalance, improve speed, and downstream generalization~\citep{roller2021hash, dua2021tricks, Thor.Zuo.2021,gross2017hard,expertchoice2022,jaszczur2021sparse}. 

\textbf{Non-uniform Architectures:} EfficientNet represents one of the very early non-uniform architectures that leverages layer heterogeneity to achieve SoTA. Instead of searching for a new operator or a new block of operators, EfficientNet focuses on optimizing the layer compound coefficients to scale the model effectively. This heterogeneity leads to a model more than 8$\times$ smaller and more than 6$\times$ faster on inference~\citep{tan2019efficientnet}. Sandwich Transformer promotes a non-interleaved, non-uniform architecture for language modeling tasks. However, the sandwich reordering pattern does not guarantee performance gains across every task. Residual MoE~\citep{wu2022residual} factorized the weights into an input-independent core and an input-dependent residual, thus achieves comparable results with the upper-bound MoE training while only introducing minor additional training cost than the lower-bound non-MoE training. In this work, we take inspiration from the earlier work but further improve scaling and generalization via automatic model discoveries.

\section{Method}
\label{sec:method}
\subsection{Deriving Our Model Components}
\label{subsec:derive-components}

\begin{figure}[!tph]
\centering\includegraphics[width=0.95\linewidth,trim={6.5cm 5cm 6.5cm, 2.5cm},clip]{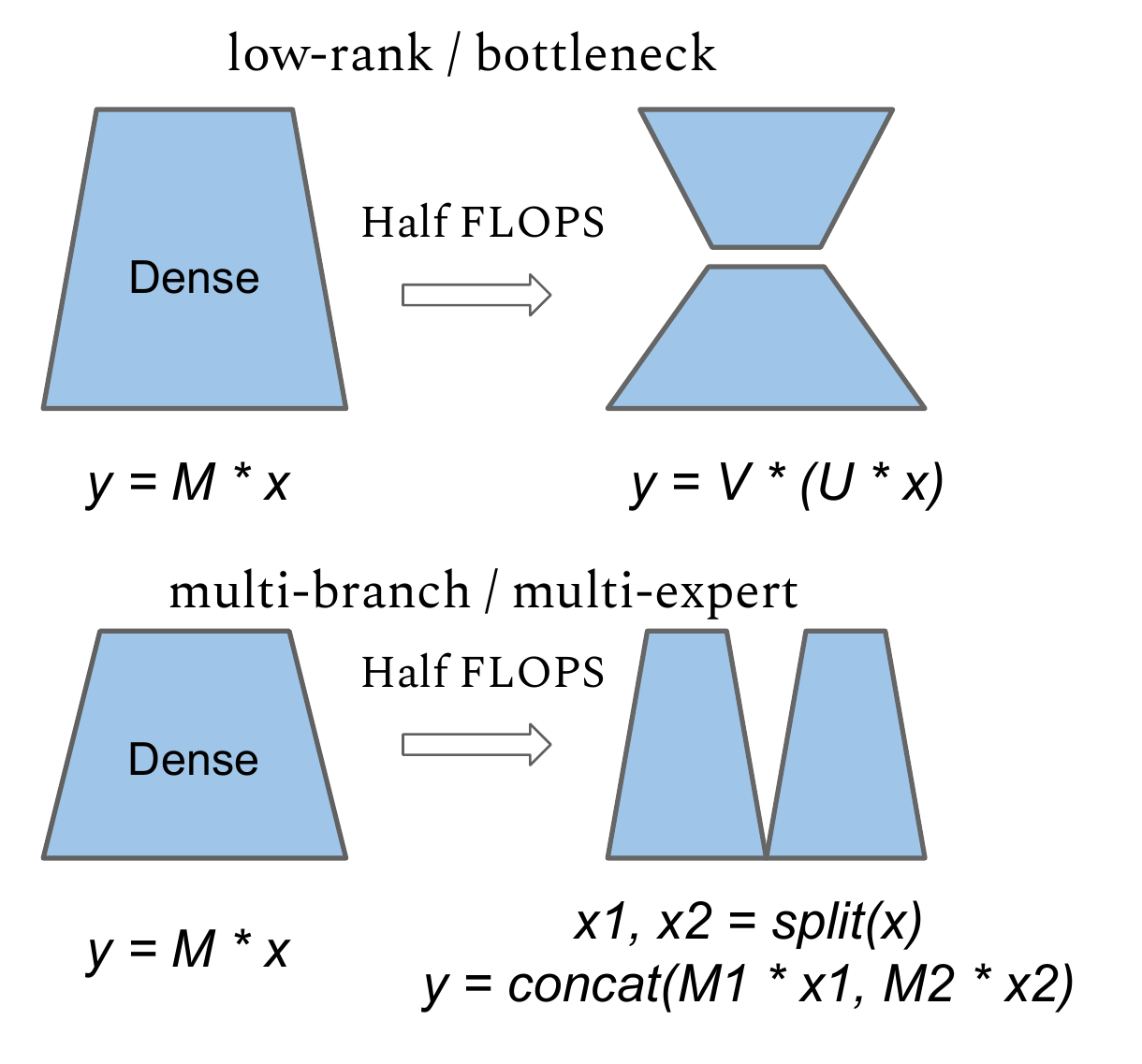}
\caption{Two methods of matrix factorization: Low-rank and Multi-branch.}\vspace{-1em}
\label{fig:derive-components}
\end{figure}

\begin{figure}[!tph]
\centering\includegraphics[width=0.9\linewidth,trim={6.5cm 1cm 5cm, 1cm},clip]{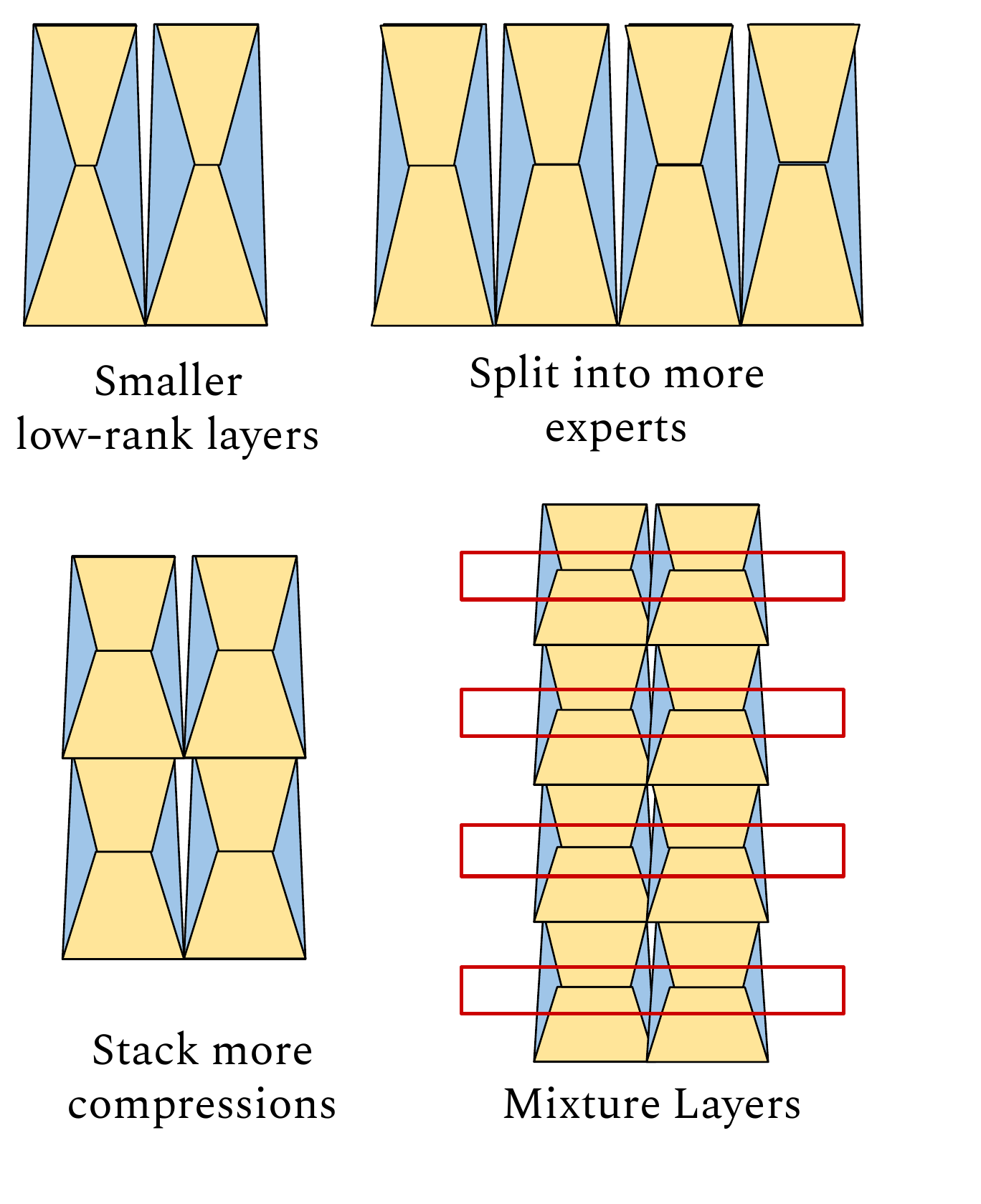}\\\vspace{-1em}
\caption{Evolving matrix factorization into transformer-styled model architecture.} \label{fig:more-compression}
\vspace{-1em}
\end{figure}

There are various forms of computation factorization that can lead to lower computation cost or faster computation without penalizing model quality. As indicated in \cref{fig:derive-components}, low-rank and multi-expert layers are two major methods for factorizing a matrix multiplication, both of which reduces FLOPs by half while not sacrificing model capacity. When devising an efficient neural network, as indicated in \cref{fig:more-compression}, low-rank and multi-expert can be combined and stacked to achieve more interesting model architectures that are computationally efficient. Finally, by also coupling a temporal mixture layer (e.g. attention~\citep{vaswani2017attention}, gMLP~\citep{liu2021pay} or MLP mixer~\citep{tolstikhin2021mlp}) which captures the causal relations between tokens, the network becomes a multi-expert transformer variant.

However, constructing an efficient network does not require conforming to the uniformity of the model architecture as illustrated in the last figure of \cref{fig:more-compression}. By carefully selecting layer types and layer interleaving, as well as other hyperparameters layers, we could achieve higher quality, training efficiency, as well as better scaling. This leads our exploration towards a more training-efficient architecture by adopting low-rank and multi-expert compression methods with coarse-grain sparsity.

\subsection{Block-wise Architecture}
\begin{table}[t]
\centering
\caption{Search Space Table: $F_\mathrm{attn}$ is a self-attention layer, $F_\mathrm{moe}$ is a sparsely gated FFN layer, and $F_\mathrm{ffn}$ is a regular dense FFN layer. The baseline is a 100M 12-layer dense transformer model with $H_\mathrm{model\_dim}$ = 768.}
\label{tab:search-space}
\begin{tabular}{ll}
\multicolumn{1}{c}{\bf Search Item}  &\multicolumn{1}{c}{\bf Search Space}
\\ \midrule 
Layer Type  ($\mathcal{F}_{i}$)  & $ \mathcal{F}_\mathrm{attn}, \mathcal{F}_\mathrm{moe}, \mathcal{F}_\mathrm{ffn} $ \\
Model Dim. ($d$) & 512, 768, 1024 \\
MoE Hidden Dim. ($d_\mathrm{moe}$) & 1536, 2048, 3072, 4096\\
FFN Hidden Dim.  ($d_\mathrm{ffn}$) & 1536, 2048, 3072, 4096 \\
Attention Heads.  ($h$) & 12, 16, 20 \\
Gating Func. ($g$)  & \textit{Top-2}, \textit{Expert Choice} \\
Capacity Factor ($c$) & 1, 2, 3, 4 \\
Activation Func. ($a$) & \textit{Gated Re/GeLU}, \textit{ReLU}, \textit{GeLU}\\
\end{tabular}

\end{table}

\begin{figure*}[!t!]
  \centering 
\includegraphics[width=0.9\linewidth,trim={0.5cm 4cm 2cm, 3cm},clip]{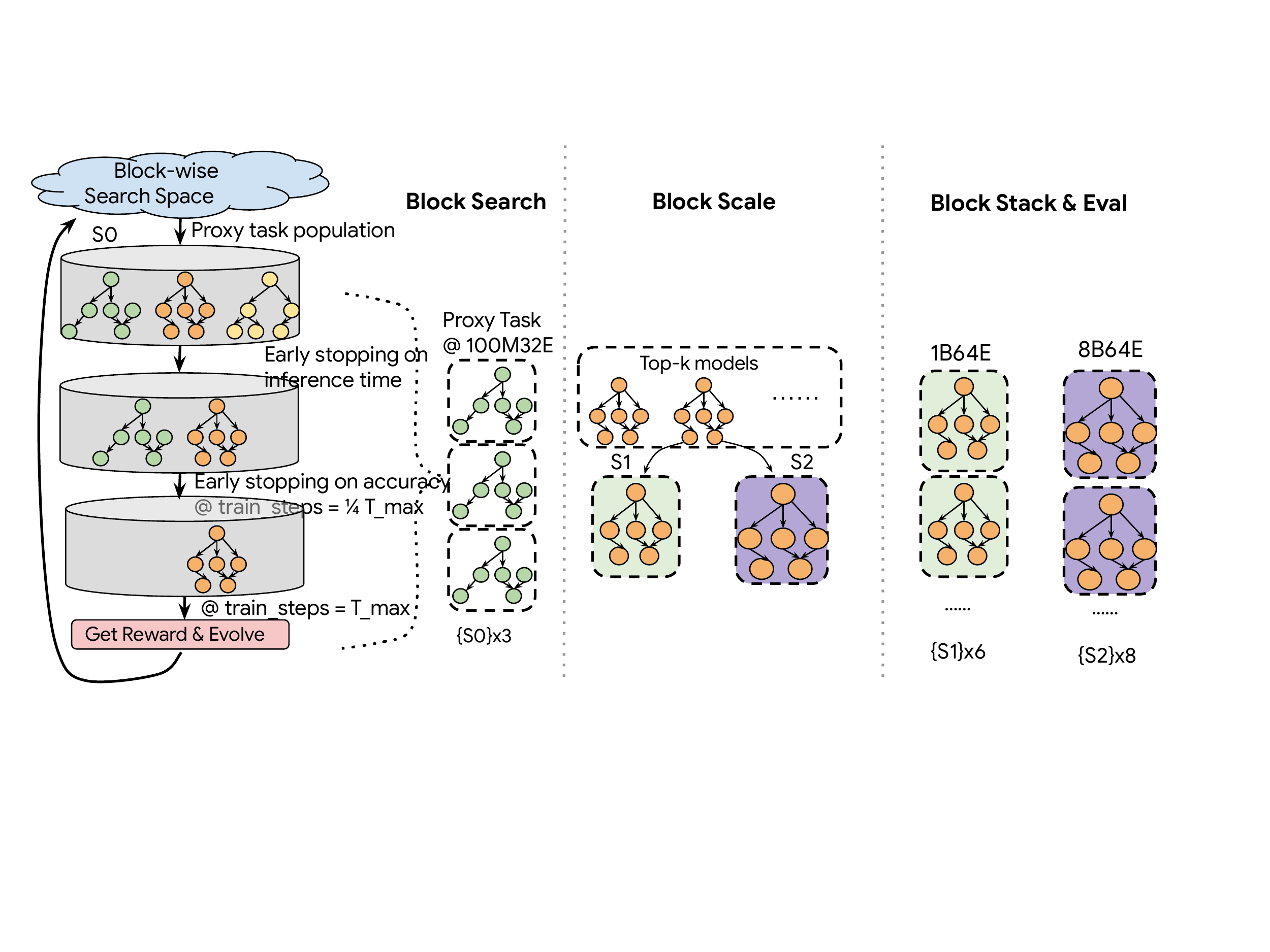}\vspace{-1em}

    \caption{Block-wise architecture search and stacking.}
    \label{fig:search}
    \vspace{-1em}
\end{figure*}

We largely take inspiration from the layer-wise compound scaling in EfficientNet~\citep{tan2019efficientnet}. For the easiness of scaling, We construct a block-wise search space where the restriction of uniformly stacking layers is removed. Instead, we create a generic layer as a function $Y_{i} = \mathcal{F}_{i}(X_{i}), \mathcal{F}_{i} \in \{\mathcal{F}_\mathrm{attn}, \mathcal{F}_\mathrm{moe}, \mathcal{F}_\mathrm{ffn}\}$ where $\mathcal{F}_{i}$ is an operator selected from the operation set consisting of self attention, sparsely gated feed-forward (MoE), and dense feed-forward sub-layers as depicted in \cref{eq:eq3}. Input $X_{i}$ has a tensor shape of $\{B, L, H\}$ and $H \in \{\frac{3}{4}, 1, \frac{3}{2}\} \times H_\mathrm{model\_dim}$ where $B$ is the batch size, $L$ is the sequence length, and $H$ is a tunable model dimension. The intuition behind tuning model dimension is to enable more flexible network topologies with various factorization methods as described in \cref{subsec:derive-components}. For example, we could instantiate a model with wider hidden dimensions or a model with experts but each expert being narrow.

Unlike a traditional simple, uniform transformer block, a Brainformer block is a complex block $\mathcal{N}$ that can be represented by a list of composed layers in~\cref{eq:eq1}: 
\begin{equation}
\label{eq:eq1}
\mathcal{N} = \mathcal{F}_{k}\odot ...\odot \mathcal{F}_{2} \odot \mathcal{F}_{1}(X_{1}) = \bigodot_{j=1...k}\mathcal{F}_{j}(X_{1})
\end{equation}

We can stack an arbitrary number of Brainformer blocks to create a target model. The search objective is to find an optimal layer architecture $\mathcal{F}_{i}$, and model scaling multipliers for multiple model inner dimensions that minimizes the perplexity. Table~\ref{tab:search-space} summarizes the search space in a Brainformer architecture. 

Figure~\ref{fig:search} and Algorithm~\ref{alg:search} illustrate the two phases that we use to discover compute-efficient Brainformer models. During the search, a regularized evolutionary search algorithm samples block architectures from the search space and trains the sampled architectures using a proxy training. In a proxy training task, a small 100M32E architecture is instantiated by stacking the sampled block three times. This matches the number of layers in a baseline GLaM architecture. We apply early stopping during the proxy training, where unpromising models are pruned early due to the violation of inference time constraint or perplexity constraint at 25\% of the maximum training steps, compared to the baseline GLaM architecture.

At the end of evolution, top-k block architectures with the highest rewards are evaluated at multiple target scales. In our evaluation, we first scale the model dimension and hidden dimension 2x and 4x, following the scaling factors presented in GLaM, to create block S1 and S2 targeting 1B and 8B model scale. Then we stack block S1 and S2 respectively to create 1B64E and 8B64E model variants. N in Algorithm~\ref{alg:search} can be determined mathematically according to the target total activated parameters. Our final evaluations are based on comparisons with baseline architectures at multiple scales.

\begin{algorithm}[h]
\caption{Brainformer Block Search} 
\label{alg:search}
\begin{algorithmic}[1]
\REQUIRE A Block-wise architecture search space $\mathcal{B}$. An evolutionary search algorithm with population size $p$.
\FOR{t = 1 to $T_0$}
    \FOR{$\mathcal{B}^{(i)}$ in $ \text{SamplePopulation}( \mathcal{B}, p)$}
        \STATE $\mathcal{G}^{(i)} \leftarrow \text{StackThreeTimes}(\mathcal{B}^{(i)})$
        \IF{$\text{EarlyStopping}(\mathcal{G}^{(i)})$}
        \STATE $\mathcal{R}^{(i)} = -1$
        \ELSE
        \STATE $\mathcal{A}^{i}, \mathcal{T}^{i} \leftarrow \text{Train}(\mathcal{G}^{(i)}, T_{max})$
        \STATE $\mathcal{R}^{(i)}  \leftarrow f(\mathcal{A}^{i}, \mathcal{T}^{i})$
        \ENDIF
    \ENDFOR
\ENDFOR
\STATE $\mathcal{G}_{topk} \leftarrow  \text{TopK}(\{\mathcal{G}^{(i)}, \mathcal{R}^{(i)}\}) $
\FOR{$\mathcal{G}^{(i)}$ in $\mathcal{G}_{topk}$}
    \STATE $\mathcal{G}^{(i)} \leftarrow \text{ScaleModelDim} (\mathcal{G}^{(i)})$
     \STATE $\mathcal{G}^{(i)} \leftarrow \text{StackNTimes} (\mathcal{G}^{(i)})$
     \STATE $\mathcal{A}^{i}, \mathcal{T}^{i} \leftarrow \text{Train}(\mathcal{G}^{(i)})$
\ENDFOR
\end{algorithmic}
\end{algorithm}

\subsection{Fair Comparisons Across Model Architectures}

Prior NLP model scaling studies~\citep{raffel2020exploring,GPT2018,GPT32020,rae2021scaling} typically explore quality scaling with fixed model capacity and training steps/tokens. For example, a scaling plot typically fixes training steps/tokens while varying the model parameters. However, when training a model, users typically have a fixed budget and can trade-off training time, compute resources, and quality to stay within that budget. If what we care about is computational cost and training convergence time, then comparing model qualities while fixing total parameters is not fair, particularly when comparing across model architectures and model families. For example, it may discriminate against models with more total parameters that consume fewer computational FLOPs, such as sparsely activated models. The GLaM paper~\citep{du2022glam} addresses this by conducting a scaling study on activated memory (which approximates the computational cost), rather than the total parameter size, on a fixed number of training tokens. However, comparing models with a fixed amount of training tokens may still also not be fair as some smaller models can benefit more from additional training data and outperform a bigger model with the same total training cost (e.g. GPU hours, TPU hours, etc.). The Chinchilla paper~\citep{hoffmann2022training} is the first to suggest compute-efficient scaling, which varies both model capacity and training tokens at a fixed computational cost. Resonating with compute-efficient model scaling, we further take model architectural change into consideration during the search for efficient model architectures with better training convergence and inference time. More particularly, we compare across models with a fixed training cost and model inference time,  which allows the search algorithm to trade off between model capacity and training tokens.  

\begin{figure}[!t!]
  \centering 
\includegraphics[width=1\linewidth,trim={2cm 7cm 4cm, 4.2cm},clip]{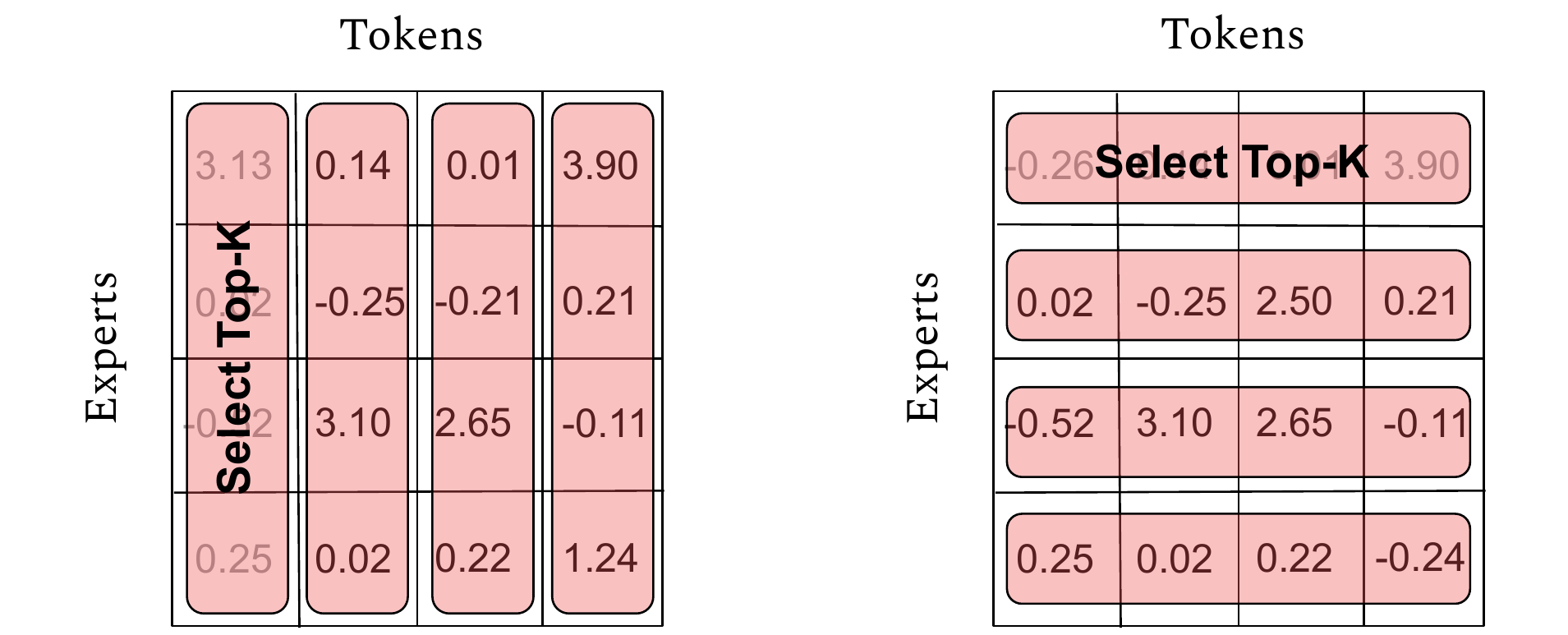}
    \caption{Token-based routing vs. Expert-based routing.}
    \label{fig:routing}
    \vspace{-1em}
\end{figure}
\begin{table*}[t]
    \centering
\small
    \caption{Sizes and architectures of baseline dense models and MoE (GLaM) models. Models are grouped by the number of activated parameters per token.
    }
    \scalebox{1.1} {
    \begin{tabular}{lccccccccc}
    \toprule 
    Model & Type &  $n_{\text{params}}$ &  $n_{\text{act-params}}$ &  $L$ & $M$ &  $H$ &  $n_{\text{heads}}$ &  $d_{\text{head}}$ &  $E$ \\
    \midrule
    0.1B & Dense & 130M & 130M &\multirow{2}{*}{12} & \multirow{2}{*}{768} & \multirow{2}{*}{3,072} & \multirow{2}{*}{12} & \multirow{2}{*}{64} & --\\
    0.1B/32E & MoE & 1.9\text{B} & 145M & & & & & & 32\\ \hline
    \midrule
    1.7B & Dense & 1.7B & 1.700B & \multirow{2}{*}{24} & \multirow{2}{*}{2,048} & \multirow{2}{*}{8,192} & \multirow{2}{*}{16} & \multirow{2}{*}{128} & --\\
    1.7B/64E & MoE & 27B & 1.879B & & & & & & 64\\ \hline
    \midrule
    
    8B & Dense & 8.7B & 8.7B &\multirow{2}{*}{32} & \multirow{2}{*}{4,096} & \multirow{2}{*}{16,384} & \multirow{2}{*}{32} & \multirow{2}{*}{128} & -\\ 
    8B/64E & MoE & 143B & 9.8B & & & & & & 64\\
    \bottomrule
    \end{tabular}}

    \label{tab:setup}
\end{table*}

\subsection{Training Time Constrained Search}
We fix the wall clock time for each search trial which encourages models with faster training convergence being discovered. The objective is to find model architectures that yield higher accuracy with a fixed training budget (number of chips times training hours). In an evolution search, a controller minimizes the pre-training validation cross-entropy loss in~\cref{eq:eq2} while meeting an inference time constraint in~\cref{eq:eq5}. The block architecture is defined around a 100M vanilla transformer architecture, as illustrated in Table~\ref{tab:setup}. Each trial is trained with a fixed wall clock time so that faster models can be compensated with more training steps. We empirically find that  fixing training wall clock time while meeting a inference time constraint yields models with faster training convergence and higher quality.
\setlength{\abovedisplayskip}{-5pt}
\setlength{\belowdisplayskip}{0pt}
\setlength{\abovedisplayshortskip}{0pt}
\setlength{\belowdisplayshortskip}{0pt}

\begin{align}
\label{eq:eq2}
\underset{\mathcal{F}_{1:k}, d, d_{\mathrm{moe}}, d_{ffn}, h, g, c, a}{\min}\mathcal{L}(\mathcal{N}(\mathcal{F}_{1:k}, d, d_{\mathrm{moe}}, d_{ffn}, h, g, c, a))
\end{align}

\begin{equation}
\label{eq:eq3}
\mathcal{F}_{i} = 
\begin{cases}
\mathcal{F}_{i}^{d, h, a}, & \text{if} \quad \mathcal{F}_{i}=\mathcal{F}_{attn}  \\

\mathcal{F}_{i}^{d, d_{ffn}, a}, & \text{else if} \quad \mathcal{F}_{i}=\mathcal{F}_{ffn}  \\

\mathcal{F}_{i}^{d, d_{\mathrm{moe}}, g, c, a}, & \text{otherwise} \quad \mathcal{F}_{i}=\mathcal{F}_{\mathrm{moe}} \\
\end{cases}
\end{equation}

\begin{align}
s.t. \quad \mathcal{N}(\mathcal{F}_{1:k}, d, d_{\mathrm{moe}}, d_{ff}, h, g, c, a) = \underset{i=1...k}{\bigodot}\mathcal{F}_{i}(X_{1})
\end{align}
 
\begin{align}
\label{eq:eq5}
    \mathrm{Step\_Time}(\mathcal{N}) \leq \mathrm{baseline\_step\_time}
\end{align}

\section{Token-based Routing Versus Expert-based Routing}
While there are various routing methods in existing MoE literature, we primarily focus on two classes of routing: token-based routing and expert-based routing, to illustrate the idea that routing strategy can change the optimal model architecture when sparsely activated layers are introduced.

As an example, in Figure~\ref{fig:routing}, the rows and columns contain un-normalized scores computed for four tokens and four experts. Each value is produced by the dot product of the token embedding and the expert embedding. Once the token-to-expert affinity scores are generated, there are a few ways to decide which experts each token should be routed to. In token-based routing, the model routes to the top-k experts for each token, while in an expert-based routing, the experts choose top-k tokens. More particularly, we follow the top-2 gating approach used in GShard~\citep{lepikhin2020gshard} and GLaM~\citep{du2022glam} as top-2 has demonstrated stronger empirical performance than top-1 gating. For the expert-based gating, we follow the Expert Choice gating~\citep{expertchoice2022} where perfect load balance is achieved with heterogeneous parameter allocation.

There are various ways of generating the token-to-expert affinity scores. One possible way is to create a trainable gating matrix $W_g$ that projects the input feature space to a token-to-expert score. The score should be normalized either along the token dimension or the expert dimension. To avoid causal leakage in decoding mode, we suggest normalizing along the expert dimension for both token-based routing and expert-based routing.

\section{Evaluation}
\label{sec:evaluation}
\begin{figure*}[!t]
\begin{tabular}{cc}
\begin{subfigure}
\centering\includegraphics[width=0.47\linewidth,trim={0cm 0cm 1cm, 1cm},clip]{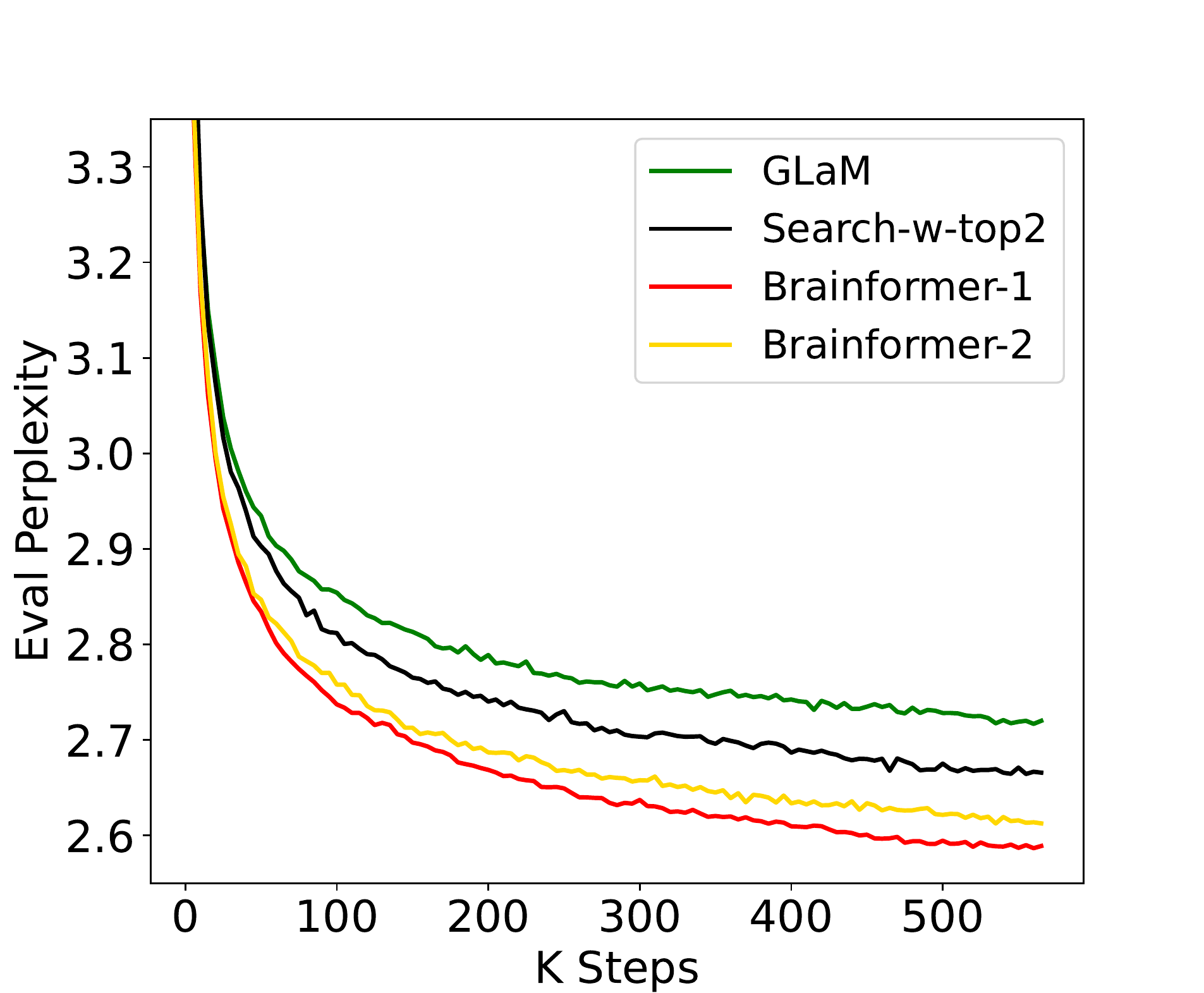}
\end{subfigure}&
\begin{subfigure}\centering\includegraphics[width=0.5\linewidth,trim={0cm 0cm 0.5cm, 1cm},clip]{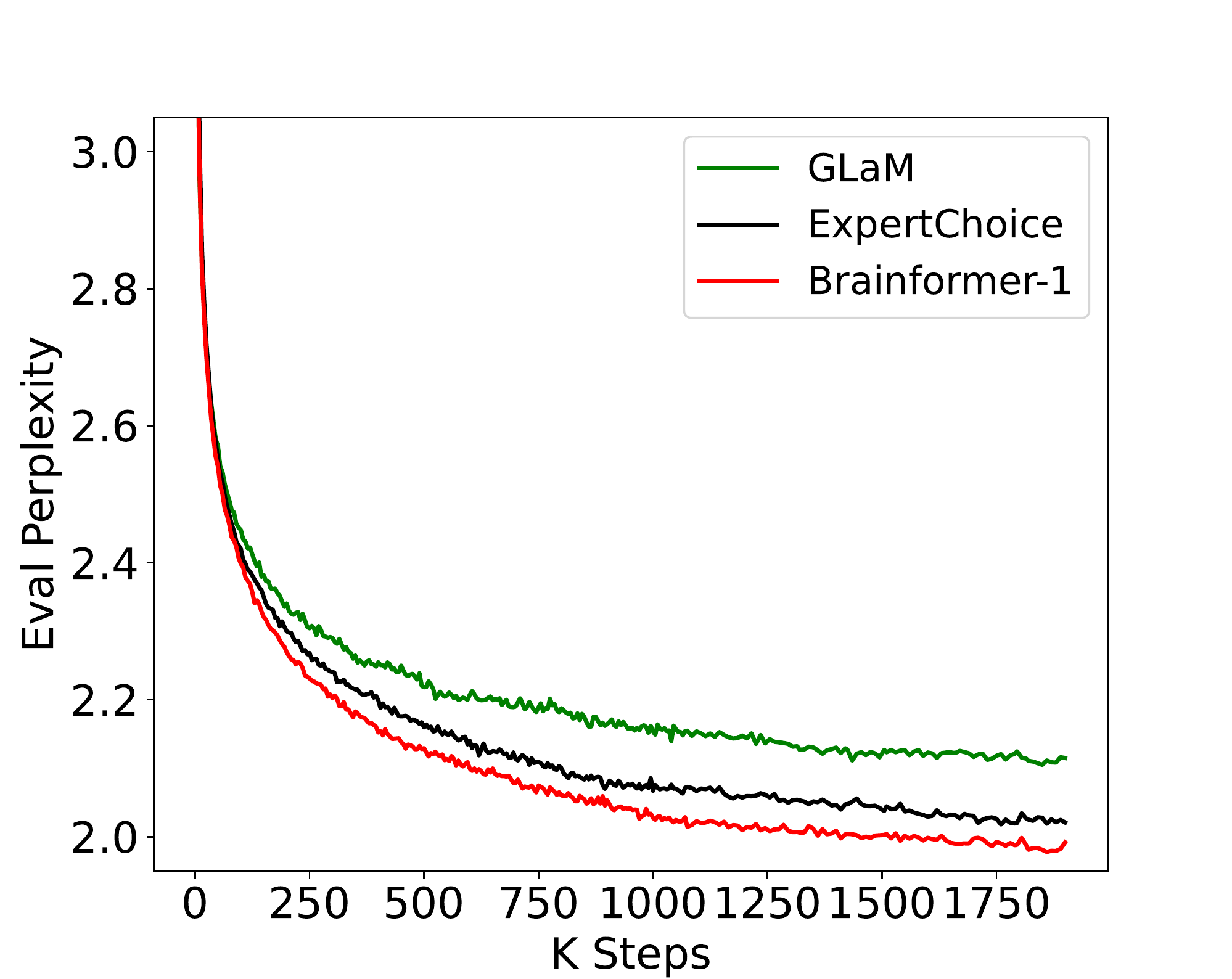}\end{subfigure}\\
(a) & (b)
\end{tabular}
\caption{(a) Pre-training perplexity comparison for 100M32E (100M parameters per expert, 32 experts). Search-w-top2 is the model found by using neural architecture search but with fixed top-2 token-based gating. (b) Training perplexity comparison for 8B64E (8B parameters per experts, 64 experts). Expert Choice is the GLaM architecture with expert-based gating function.}
\label{fig:perplexity}
\vspace{-1em}
\end{figure*}

\textbf{Setup:} Table~\ref{tab:setup} summarizes the hyperparameter settings of different baseline MoE models. In the baseline MoE GLaM~\citep{du2022glam} model, we interleave transformer blocks with regular dense FFNs and transformer blocks with sparsely gated FFNs (MoE layer). As a reference point, we also include the respective dense model configurations with comparable numbers of activated parameters per-token during inference in the table. With a similar number of activated parameters as a 0.1B dense model, 0.1B/32E represents the sparse model with every other transformer layer replaced by a 32-expert MoE layer. While $n_{\mathrm{params}}$ is the total number of trainable parameters, $n_{\mathrm{act-params}}$ represents the number of activated parameters per token. $n_{\mathrm{act-params}}$ roughly approximates the computational expensive of a model. $L$ is the total number of Transformer layers, $M$ is the model dimension, $H$ is the hidden dimension after the projection in each transformer layer, $n_{\mathrm{heads}}$ is the number of attention heads, and $d_{\mathrm{head}}$ is the hidden dimension of each attention head. We train and evaluate our Brainformer models and baseline models on 64 Cloud TPU-V4 chips, except for models at the 8B-scale which take 512 Cloud TPU-V4 chips to train. 

\begin{table*}[!htp]
\centering
\caption{Training efficiency comparison. Brainformer models have better training convergence and faster step times, compared to GLaM, fixed gating search, and expert-based gating but with fixed architecture. Brainformer-1 and Brainformer-2 are two selected best models. With limited computational resources, we only scale Brainformer-1 to 1B and 8B scales.}
\vspace{0.5em}
\resizebox{1.8\columnwidth}{!}{
\begin{tabular}{ccccccc}
\multicolumn{1}{c}{\bf Model}  &\multicolumn{1}{c}{\bf Total Params} &\multicolumn{1}{c}{\bf Activated Params} &\multicolumn{1}{c}{\bf Train Steps} &\multicolumn{1}{c}{\bf Steps/Sec} &\multicolumn{1}{c}{\bf PPLX} 
\\ \toprule
\textbf{100M32E} &&&&&& \\
GLaM & 1B & 145M & 0.5M & 1.92 & 2.73 +/- 0.002 &  \\ 
Search-w-Top2 & 1.87B & 210M & 0.5M & 2.03 & 2.67 +/- 0.005 & \\
Brainformer-1 & 3.19B & 156M & 0.5M & 2.03 & \textbf{2.57} +/- 0.003 & \\
Brainformer-2 & 3.33B & 266M & 0.5M & \textbf{2.16} & 2.59 +/- 0.005 & \\
\midrule
\textbf{1B64E} &&&&&& \\
GLaM & 27B & 1.88B & 1.0M & 1.23 & 2.25 +/- 0.004 &  \\ 
Search-w-Top2 & 27B & 3.05B & 1.0M & 1.27 & \textbf{2.21} +/- 0.003 & \\
Brainformer-1 & 30B & 1.38B & 1.0M & \textbf{2.00} & 2.25 +/- 0.002 & \\
Brainformer-2 & 52B & \textbf{1.31B} & 1.0M & 1.76 & 2.23 +/- 0.001 & \\
\midrule
\textbf{8B64E} &&&&&& \\
GLaM & 143B & 9.8B & 1.5M & 0.39 & 2.12 +/- 0.002 &  \\ 
Expert-based Gating & 143B & 9.8B & 1.5M & 0.50 & 2.03 +/- 0.005 & \\
Brainformer-1 & 158B & \textbf{7.4B} & 1.5M & \textbf{1.96} & \textbf{1.99} +/- 0.002 & \\
\bottomrule

\end{tabular}
}

\label{tab:train-efficiency}
\end{table*}

\textbf{Dataset:} We use the high-quality dataset from GLaM of 1.6 trillion tokens that are representative of a wide range of natural language use cases. This dataset consists of a high-quality filtered subset of webpages that are combined with smaller corpora of books, Wikipedia pages, conversations, forums, and news to create the final dataset. A more detailed description of the dataset including the data and mixture weights can be found in the GLaM paper~\citep{du2022glam}.

\textbf{Model Training:} We train a few decoder-only models using the searched best Brainformer blocks and related baselines. Brainformer-1 and Brainformer-2 are two selected best models. With limited computational resources, we only scale Brainformer-1 to 1B and 8B scales. Our model training follows the setup of GLaM where a maximum sequence length of 1024 tokens is used. We use an Adafactor optimizer~\citep{shazeer2018adafactor} with first-moment decay $\beta_{1}=0$ and second-moment decay $\beta_{2}=0.99$. The learning rate is kept constant for the first 10K training steps, then is  decayed with an inverse square root schedule. We use the SentencePiece subword tokenizer with a vocabulary of size of 256K. The 100M-scale models and 1B-scale models are trained with 64 TPU V4 chips, while the largest model (8B/64E) evaluated is trained on 512 TPU V4 chips. 
We don't use any dropout during training because the training corpus is large enough that each sample is only encountered once.

\textbf{Model Evaluation:} We mainly focus on two types of downstream evaluation: 1) Fine-tuning performance on 11 selected classification tasks from the GLUE and SuperGLUE benchmarks~\citep{wang2019glue, wang2020superglue}. 2) We evaluate oneshot performance with five language generation tasks focused on question answering.

\begin{table*}[!htp]
\centering
\caption{Finetuning Results on GLUE/superGLUE: Brainformers at 100M and 1B significantly outperform GLaM counterparts, yielding over 3\% gains in overall scores.}
\vspace{0.5em}
\begin{tabular}{cccccccc}
\multicolumn{1}{c}{\bf Size} 
&\multicolumn{1}{c}{\bf Model} &\multicolumn{1}{c}{\bf BoolQ} 
&\multicolumn{1}{c}{\bf CB} & \multicolumn{1}{c}{\bf CoLA} &\multicolumn{1}{c}{\bf MNLI} &\multicolumn{1}{c}{\bf MRPC} &\multicolumn{1}{c}{\bf QNLI} \\
\toprule
\multirow{2}{*}{100M64E} & GLaM & 0.791 & 0.859 & 0.818 & 0.849 & 0.833 & 0.901 \\ 
& Brainformer-1 & 0.812 & 0.922 & 0.828 & 0.855 & 0.870 & 0.907 \\ 
\midrule
\multirow{2}{*}{1B64E} & GLaM & 0.829 & 0.938 & 0.831 & 0.860 & 0.857 & 0.919 \\ 
& Brainformer-1 & 0.859 & 0.938 & 0.863 & 0.896 & 0.875 & 0.938 \\
\midrule
\multicolumn{1}{c}{\bf Size} 
&\multicolumn{1}{c}{\bf Model} &\multicolumn{1}{c}{\bf QQP} &\multicolumn{1}{c}{\bf RTE} &\multicolumn{1}{c}{\bf SST2} &\multicolumn{1}{c}{\bf WiC} & \multicolumn{1}{c}{\bf WNLI} &\multicolumn{1}{c}{\bf AVG} \\
\midrule
\multirow{2}{*}{100M64E} & GLaM & 0.907 & 0.808 & 0.952 & 0.687 & 0.609 & 0.819 \\ 
& Brainformer-1 & 0.812 & 0.840 & 0.952 & 0.702 & 0.635 & \textbf{0.840} \\
\midrule
\multirow{2}{*}{1B64E} & GLaM & 0.911 & 0.816 & 0.945 & 0.711 & 0.547 & 0.833  \\ 
& Brainformer-1 & 0.917 & 0.899 & 0.972 & 0.720 & 0.719 & \textbf{0.873} \\
\bottomrule
\end{tabular}
\label{tab:finetuning}
\end{table*}

\subsection{Training Convergence}
\vspace{-0.5em}
In this section, we evaluate Brainformer top models with related baselines including 1) Top-2 gating based model architecture search (Search-w-Top2) and 2) GLaM~\citep{du2022glam}, a manually crafted architecture with fixed top-2 gating. Providing the flexibility of tuning the gating function and network architecture significantly improves pre-training efficiency. As shown in \cref{tab:train-efficiency}, our searched best Brainformer models outperform the baselines in terms of computational cost (activated parameters), training step time (steps/sec), and training perplexity (PPLX) for fixed training steps. When scaled to 8B64E, Brainformer converges to lower perplexity and is more than 5x faster in step time and 2x faster in training convergence using the same hardware configuration (512 Cloud TPU-V4 chips). With a fixed 600B training tokens, Brainformer is much more accurate than the baselines at 8B scale.

\subsection{Finetuning Results}
We pretrain the models for a total \textbf{fixed wall clock time} as the baseline GLaM model. We then finetune the models with eleven selected GLUE and SuperGLUE classification tasks. At two different scales, 100M64E and 1B64E, Brainformers outperform the baseline GLaM model by a significant margin of 2-4\% average score. The fine-tuning results in ~\cref{tab:finetuning} indicates that Brainformer not only excels at training convergence but also generalizes well to downstream tasks. 

\subsection{Fewshot Results}
\begin{table*}[!pt]
\centering
\caption{Oneshot evaluation on five important generative tasks. All models are trained with 200B training tokens.}
\vspace{0.5em}
\begin{tabular}{cccccccc}
\multicolumn{1}{c}{\bf Model}  &\multicolumn{1}{c}{\bf Nqs} &\multicolumn{1}{c}{\bf Triviaqa} &\multicolumn{1}{c}{\bf Webqa} &\multicolumn{1}{c}{\bf Squadv2} & \multicolumn{1}{c}{\bf Lambada} & \multicolumn{1}{c}{\bf Steps/Sec} 
\\ \toprule

GLaM 1B64E & \textbf{9.14} & 41.8 & 10.8 & 46.2 & 25.2 & 0.55 \\ 
Primer 1B~\citep{so2021searching} & 4.82 & 24.7 & 6.50 & 49.2 & 22.6 & 1.50 \\  
Brainformer 1B64E & 8.23 & \textbf{43.4} & \textbf{12.0} & \textbf{49.5} & \textbf{25.7} & 1.37 \\
\bottomrule
\end{tabular}
\vspace{-1em}
\label{tab:fewshot}
\end{table*}

Aligned with prior work in fewshot in-context learning, we compare Brainformer oneshot performance on five selected generative tasks in~\cref{tab:fewshot}: Natural Questions~\citep{NQS}, TriviaQA~\citep{JoshiTriviaQA2017}, Web Questions~\citep{Webqa2013}, Squadv2~\citep{squad2018}, and Lambada~\citep{lambada}, with a sparse model GLaM and a dense model Primer~\citep{so2021searching} of similar activated memory size. Brainformer outperforms Primer and GLaM by a large margin on all the tasks except Nqs being slightly worse than GLaM. GLaM yields competitive scores while being 2x slower than Brainformer.

\section{Discussion}
\vspace{-0.5em}

\subsection{Visualizing a Brainformer Block}
In this section, \cref{fig:brainformer_block} provides a visualization of a Brainformer architecture block. Unlike a conventional transformer block, where there is only an attention layer and a dense feed-forward layer, a Brainformer block contains 8 sub-layers. The Brianformer block is repeated 3 times, 6 times, and 8 times respectively in the 100M, 1B, and 8B scale. In a vanilla transformer model, a dense FFN layer has an optimized expansion ratio of 4, which results in a hidden dimension 4x wider than the model dimension. In the optimized Brainformer block 1 and 2, the search algorithm picks a slightly larger model dimension of 1024 (as compared to 768) and a smaller expansion factor in the dense FFNs and MoE layers (as compared to 3072). This is a reasonable optimization, as MoE layers effectively widen the network with more experts. In the MoE layers, the search algorithm picks the expert choice gating function~\citep{expertchoice2022} with a capacity factor of one in Brainformer block 1, resulting in a very sparse network in which each token can be routed to a single expert on average. Being much faster in step time, block 1 takes more training steps, thus training data to achieve good quality. Therefore, we also picked another strong candidate, Brainformer block 2, in which a larger capacity factor in the MoE layers is selected. Block 2 is lightly slower in step time, but takes fewer training steps to get good accuracy, thus is more data efficient.

\subsection{Can We Simplify?}
We did an ablation study on block simplification. A very natural question to ask is whether we can simplify the architecture block. In exploring the answer to this question we were able to extrapolate some patterns. We find that the ratio of different layer types is critical to model quality: replacing a layer with a different layer results in degraded quality. However, the network is relatively insensitive to layer order, such that swapping any two layers would not affect performance much. For example, to create a simplified pattern, we can interleave the dense FFNs and MoE layers or simply creating contiguous layers of the same type.
\begin{figure}[!thp]

\centering 
\includegraphics[width=0.65\linewidth,trim={4cm 3cm 4cm, 2cm},clip]{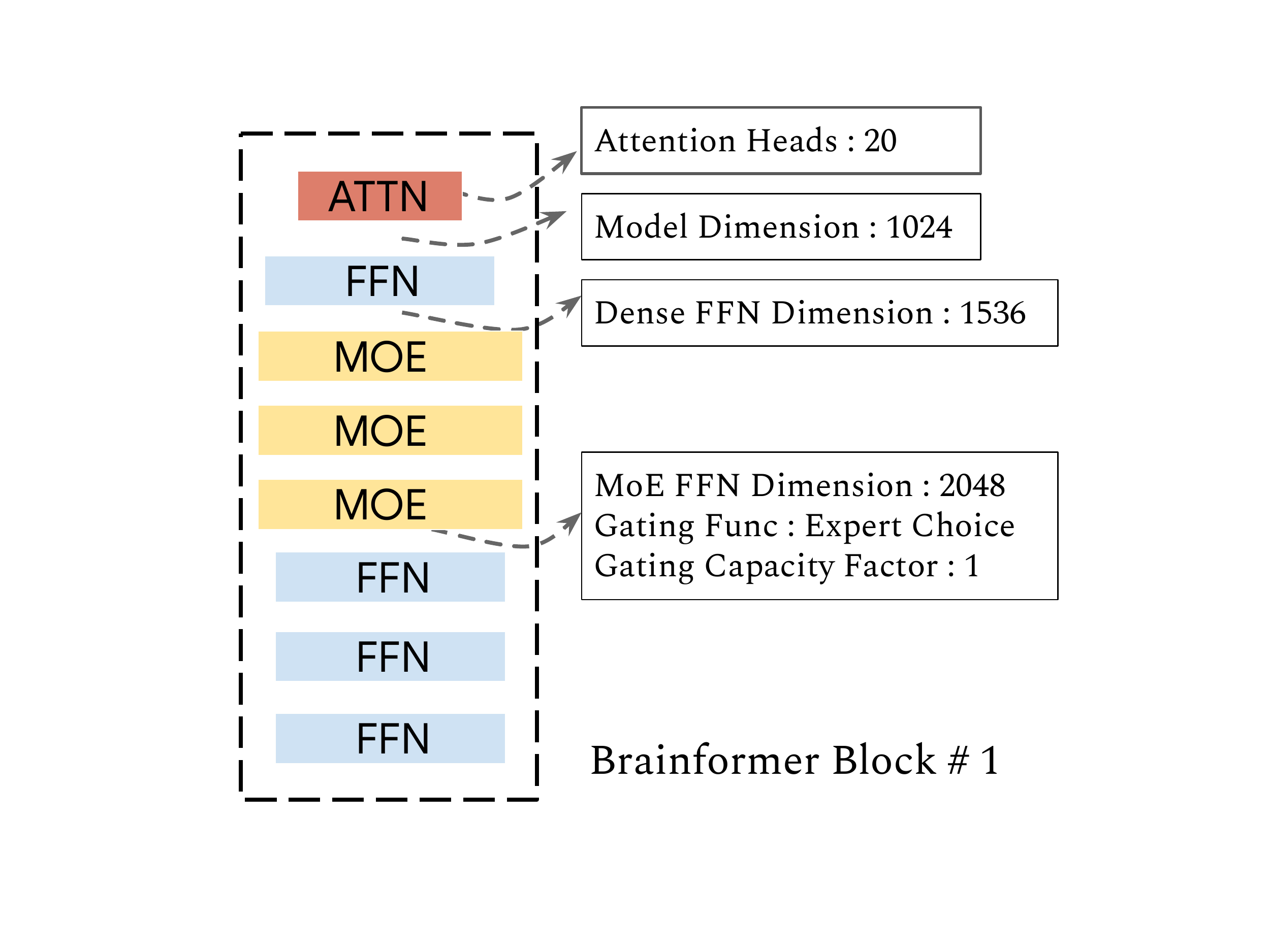}
    \caption{Brainformer Block \# 1}
    \label{fig:brainformer_block}
    \vspace{-1em}
\end{figure}
\begin{figure}[!thp]

\centering 
\includegraphics[width=0.65\linewidth,trim={4cm 3cm 4cm, 2cm},clip]{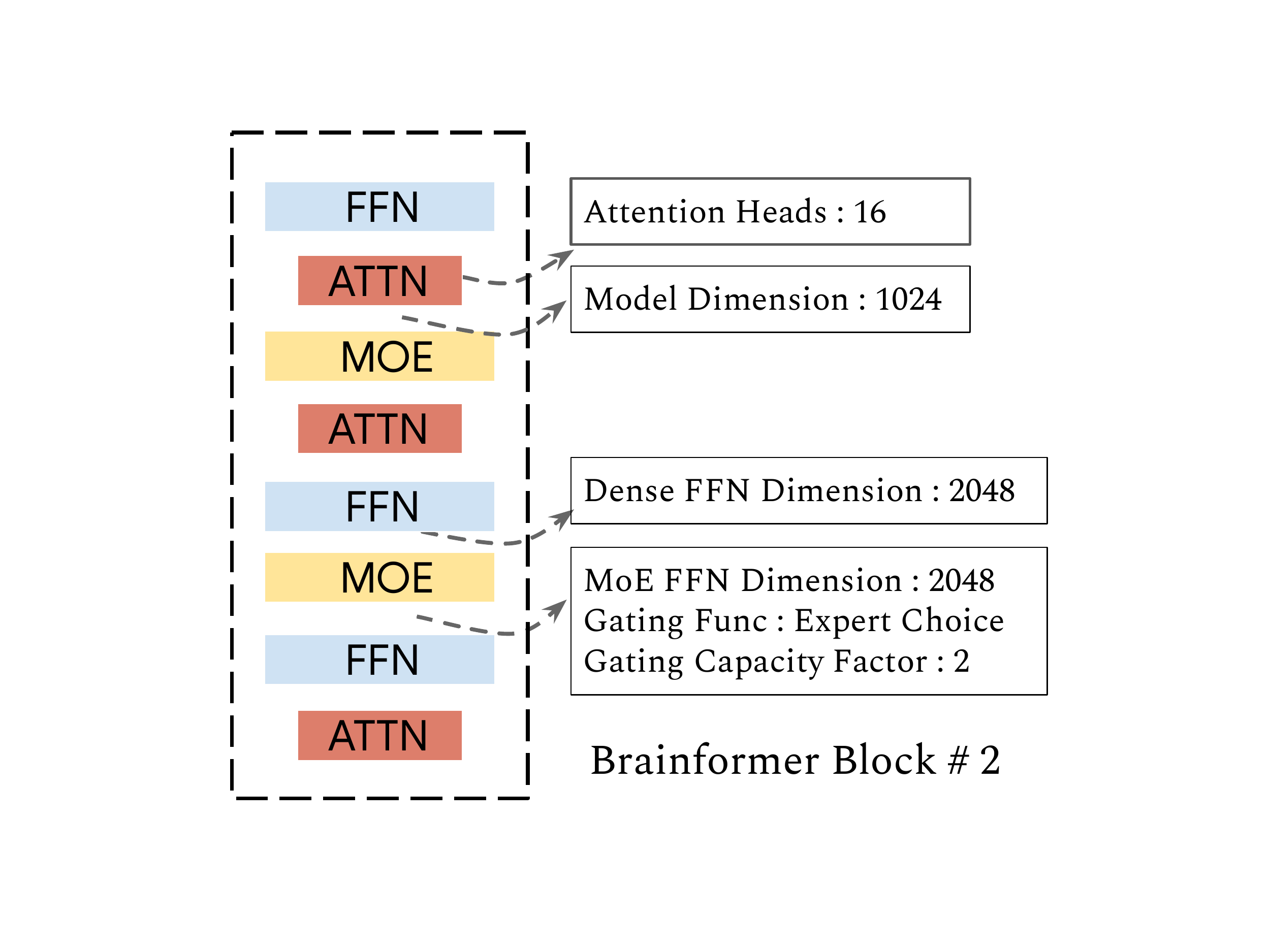}
    \caption{Brainformer Block \# 2}
    \label{fig:brainformer_block}
    \vspace{-1em}
\end{figure}

\section{Conclusion}
Using an evolutionary search algorithm, we have developed and evaluated a complex architecture block, named Brainformer, that consists of a diverse sequence of layers, including a sparsely gated feed-forward layer. Along with the new block, we also propose evaluating using a fixed training time search, which enables fair comparisons across model families. Brainformer demonstrates up to 2$\times$ faster training convergence and 5$\times$ faster step time compared to its GLaM counterpart. In downstream task evaluation, Brainformer also demonstrates a 3\% higher SuperGLUE score with fine-tuning compared to GLaM, and greatly outperforms Primer on oneshot evaluation for five generative tasks. 

\section{Limitations}
In terms of research scope, our empirical results are primarily on NLP domain, thoroughly on a wide range of NLU and NLG tasks. However, we leave it to future work to apply Brainformer to computer vision.

When adopting Brainformer targeting different hardware platforms, there can be potential intricacies. For example, edge devices can impose strict hardware constraints that restricts the expression of Brainformer models. 
A practical way is to run model training and quality evaluation on faster accelerators such as GPUs or TPUs while simulating the step time for the target hardware or using a learnt performance model to predict the inference speed on the target hardware. Another issue is some fundamental operators might not be supported on a device lacking sufficient on-chip memories. For example, global pooling is not supported on edge TPU. But that can be out of scope for this paper, as Brainformer aims to construct a compute-efficient model architecture out of feasible operators.

Another limitation can be large resource consumption. In the Brainformer search, we used 512 TPU v4 for a week to arrive at the best solutions. However, worth mentioning that we are working at a much large model scale and this will be mitigated when we use a smaller model size and smaller number of experts in the MoE layers. Also, the search identified better model architecture within as early as 500 trials. Practically, the resource consumption can be small if we only need to identify better but suboptimal models.



\nocite{langley00}

\bibliography{example_paper}
\bibliographystyle{icml2023}

\newpage
\appendix
\onecolumn
\section{You \emph{can} have an appendix here.}

You can have as much text here as you want. The main body must be at most $8$ pages long.
For the final version, one more page can be added.
If you want, you can use an appendix like this one, even using the one-column format.

\end{document}